%% file: paper.tex
\documentclass[runningheads]{llncs}
\usepackage[T1]{fontenc}
\usepackage{graphicx}
\usepackage{booktabs}
\usepackage[misc]{ifsym}
\newcommand{\corr}{(\Letter)}

\usepackage{amsmath} 
\usepackage{amssymb} 
\usepackage{multirow} 

\usepackage{pifont}
\newcommand{\cmark}{\ding{51}}
\newcommand{\xmark}{\ding{55}}
\newcommand{\tablespace}{\hspace{0.02\textwidth}}

\begin{document}

\title{SPATA: Systematic Pattern Analysis for Detailed and Transparent Data Cards}

\titlerunning{SPATA: Systematic Pattern Analysis}

\author{Jo{\~{a}}o Vitorino\inst{1,2}\orcidID{0000-0002-4968-3653} \corr \and
Eva Maia\inst{1}\orcidID{0000-0002-8075-531X} \and
Isabel Pra{\c{c}}a\inst{1}\orcidID{0000-0002-2519-9859} \and
Carlos Soares\inst{2}\orcidID{0000-0003-4549-8917}}


\authorrunning{J. Vitorino et al.}

\institute{GECAD, ISEP, Polytechnic of Porto \email{jpmvo@isep.ipp.pt}
\and
Faculty of Engineering, University of Porto}

\maketitle

\begin{abstract}
Due to the susceptibility of Artificial Intelligence (AI) to data perturbations and adversarial examples, it is crucial to perform a thorough robustness evaluation before any Machine Learning (ML) model is deployed. However, examining a model’s decision boundaries and identifying potential vulnerabilities typically requires access to the training and testing datasets, which may pose risks to data privacy and confidentiality. To improve transparency in organizations that handle confidential data or manage critical infrastructure, it is essential to allow external verification and validation of AI without the disclosure of private datasets. This paper presents Systematic Pattern Analysis (SPATA), a deterministic method that converts any tabular dataset to a domain-independent representation of its statistical patterns, to provide more detailed and transparent data cards. SPATA computes the projection of each data instance into a discrete space where they can be analyzed and compared, without risking data leakage. These projected datasets can be reliably used for the evaluation of how different features affect ML model robustness and for the generation of interpretable explanations of their behavior, contributing to more trustworthy AI.

\keywords{Data card \and Interpretability \and Robustness \and Privacy \and Machine learning \and Deep learning}
\end{abstract}

\setcounter{footnote}{0}

\input{section1}

\input{section2}

\input{section3}

\input{section4}

\input{section5}

\begin{credits}
\subsubsection{\discintname}
The authors have no competing interests to declare that are
relevant to the work presented in this paper.
\end{credits}

\bibliographystyle{splncs04}
\bibliography{thebibliography}

\end{document}

%% file: section1.tex
\section{Introduction}
\label{sec:section1}

As the use of Machine Learning (ML) models for classification tasks continues to grow, their lack of robustness is becoming a major security concern, especially for organizations that handle confidential data or manage critical infrastructure~\cite{Jedrzejewski2024}. ML models can have very different architectures, from decision tree ensembles~\cite{Chen2021} to more complex artificial neural networks~\cite{Ozkan2024}, but they are all inherently vulnerable to adversarial examples: data instances with small perturbations that exploit hidden flaws in a model's decision boundaries~\cite{Tocchetti2025}.

Due to the threat posed by adversarial examples, it becomes crucial to perform a thorough and transparent robustness assessment before deploying any ML model, to ensure that an organization understands its behavior and can explain its misclassifications~\cite{Chander2025}. Furthermore, as novel legislation such as the European Union Artificial Intelligence (AI) Act~\cite{EUAIAct} begins defining security and governance requirements, organizations must start disclosing data cards with information about the utilized datasets, and allow external verification and validation of the behavior of their models~\cite{Panigutti2023}.

However, to generate useful adversarial examples and use them to examine a model's decision boundaries, it is typically necessary to access and analyze at least a portion of the training and testing datasets~\cite{VitorinoSoK}. If that dataset contains sensitive data or proprietary information, an organization cannot share it to benefit from an external robustness assessment, as it would put data privacy and confidentiality at risk. Therefore, it is essential to enable statistical analyses and adversarial example generation in a way that helps identify potential vulnerabilities without exposing private datasets.

This paper presents Systematic Pattern Analysis (SPATA), a deterministic method that converts any tabular dataset to a domain-independent representation of its statistical patterns, to provide more detailed and transparent data cards. SPATA computes the projection of each data instance into a discrete space where they can be analyzed and compared, creating projected datasets to eliminate the need to use the original data. This method aims to contribute to the trustworthiness of AI by enabling the evaluation of how different features affect ML model robustness and the generation of interpretable explanations of their behavior, while maintaining data privacy and confidentiality. The main contributions of this paper can be summarized in three key points:

\begin{enumerate}
    \item A novel method, SPATA, for the projection of tabular datasets into a domain-independent space, performing the discretization of each feature into hierarchical bin numbers that represent a specific subdomain of that feature, followed by the systematization of the statistical patterns across multiple data instances as the possible combinations of feature subdomains.
    
    \item An optimized open-source implementation~\footnote{\url{https://github.com/vitorinojoao/spata}} to automatically analyze, export, and visualize the patterns of any given tabular dataset, which is available in the Python 3 programming language and relies on computationally efficient C language functions.
    
    \item An experimental validation of the method's reliability by creating projections of well-established datasets of the cybersecurity field, and comparing the generalization, robustness, and feature importances of ML classification models across the original and projected datasets.
\end{enumerate}

This paper is organized into multiple sections. Section 2 provides an overview of previous work on the systematization of datasets and models, focused on the evaluation of ML robustness. Section 3 describes the SPATA method with illustrative examples, and Section 4 presents an experimental validation. Finally, Section 5 provides the main conclusions and future research directions.

%% file: section2.tex
\section{Related Work}
\label{sec:section2}

In recent years, there has been progress in ML robustness, with different adversarial defense strategies being developed to safeguard ML models against data perturbations~\cite{Zhou2022}. However, despite the efforts of both the industry and the scientific community, there is still a lack of approaches to reliably assess robustness when a dataset cannot be directly accessed and analyzed~\cite{Chander2025}.

Even though adversarial examples are typically associated with attacks that specifically craft them to deceive a model, they can also occur naturally due to generalization flaws during a model's training process~\cite{Stutz2019}. This is especially concerning for organizations in the cybersecurity sector, where the integration of ML in security systems improved their capabilities, but also created more vulnerabilities~\cite{Salem2024}. For instance, when ML models started being used in intelligent security systems to detect network cyber-attacks, attackers started targeting the models themselves, performing adversarial attacks that attempt to exploit the generalization flaws to evade detection~\cite{Rosenberg2021}.

These flaws often begin in the datasets themselves, being caused by lack of data diversity or inconsistencies in the training data~\cite{Hendrycks2021}. Existing approaches to create dataset documentation in the form of data cards typically summarize their main characteristics, such as feature types, descriptions, and class labels, but often overlook the underlying patterns and data distribution of each class~\cite{Pushkarna2022}. When a model is trained without an analysis of a dataset's inconsistencies and potential biases, it may learn incorrect simplifications of its target domain, so it is necessary to document how each feature may impact robustness~\cite{Ilyas2019}.

Furthermore, while there are reporting mechanisms for models' performance, they often lack transparency into the reasoning behind their predictions, which is necessary to be able to identify possible vulnerabilities~\cite{Crisan2022}. Since the decision boundaries of ML models are not being fully analyzed, it is difficult to correctly assess and compare how robust different types of models actually are across different datasets~\cite{Tocchetti2025}. There are ongoing efforts to create better data cards to prevent data flaws from affecting a model's performance, but they still do not provide enough information for robustness and explainability~\cite{Donald2023}.

Overall, to the best of our knowledge, the current approaches for systematizing dataset characteristics and creating data cards do not eliminate the need to analyze private data, as not enough information is provided for statistical analyses and for adversarial example generation for ML robustness assessments.

%% file: section3.tex
\section{SPATA Method}
\label{sec:section3}

SPATA is based on the concept of a domain-independent space, where different data instances are represented in a deterministic manner, regardless of their values in the original domain. It was developed to enable the analysis and comparison of multiple datasets without the need to directly access private data.

The method creates dynamic and consistent representations for different data instances by analyzing the statistical patterns of each feature of a tabular dataset, both individually and in a combined manner. This facilitates the creation of data cards and the verification and validation of ML model behavior, as the projected datasets can be used to evaluate how different features affect the robustness of a model as well as to generate interpretable explanations of its behavior, without requiring access to the original data.

An optimized open-source implementation of the method is available to automatically analyze the patterns of any given tabular dataset, performing a discretization of each feature and a systematization of their underlying patterns. The implementation is available in the Python 3 programming language and relies on C language functions of the \textit{numpy} library \cite{Numpy2020} for efficient numerical computations and vectorized array operations. The patterns can be exported to the JSON and CSV formats to be reutilized, and they can also be visualized in an interactive graphical form that relies on the \textit{matplotlib} library \cite{Matplotlib2007}.

The following subsections provide the formalization of the method, divided in the definition of a feature domain and subdomains, the mapping of an original value to its projection, and the analysis performed with that domain-independent representation. Each subsection includes illustrative examples of how the method can be applied, considering three different features.

\subsection{Feature Subdomains}

To achieve a domain-independent representation of a feature, the first step is extracting useful knowledge from its own domain, through an analysis of its statistical properties and the identification of patterns in its distribution of values.

Given a tabular dataset \begin{math} X \end{math} with \begin{math} n \end{math} rows and \begin{math} m \end{math} columns, and considering \begin{math} 0 \leq i < n \end{math}, a row at index \begin{math} i \end{math} is a data instance denoted by \begin{math} X_{i*} \end{math}. Likewise, considering \begin{math} 0 \leq j < m \end{math}, a column at index \begin{math} j \end{math} is a feature vector denoted by \begin{math} X_{*j} \end{math}.

The domain of a feature can be considered as the set of all its possible values. In a continuous floating-point feature, it is the set of all real values between the minimum and maximum values of the feature vector \begin{math} X_{*j} \end{math} and can be defined as:

\begin{equation}
    D \left( X_{*j} \right) \ = \ \big\{ \ x \in \mathbb{R} :
    \ {min} \left( X_{*j} \right)
    \ \leq \ x \ \leq
    \ {max} \left( X_{*j} \right)
    \ \big\}
\end{equation}

By analyzing how a feature's values are distributed across its domain, it can be discretized into multiple subdomains that take into account the variability of that feature. Let \begin{math} \mu \left( X_{*j} \right) \end{math} be the arithmetic mean and \begin{math} \sigma \left( X_{*j} \right) \end{math} be the population standard deviation of \begin{math} X_{*j} \end{math}. These two values are used as the basis of the discretization to adapt the subdomains to the "center" of the distribution of values of a feature and to how much those values are "spread out".

A subdomain of a continuous floating-point feature at index \begin{math} j \end{math} is the set of all real values in a specific interval, calculated according to \begin{math} \mu \left( X_{*j} \right) \end{math} and \begin{math} \sigma \left( X_{*j} \right) \end{math}, that remain within the minimum and maximum values of \begin{math} X_{*j} \end{math}. Each subdomain corresponds to a specific bin number \begin{math} s \end{math}, ranging from 1 to a chosen maximum number of bins \begin{math} b \end{math}. The chosen \begin{math} b \end{math} is required to be an odd number higher or equal to 3, to enable the use of the center bin number \begin{math} b' = \left( b + 1 \right) / 2 \end{math} to represent the bin that contains the mean of a feature and all the values close to it, differentiating it from the other bins. Therefore, a feature subdomain can be defined as:

\begin{equation}
    \begin{aligned}
        &D^s \left( X_{*j}, b \right) \ = \
        \\[1.2ex]
        &\begin{cases}
            \begin{aligned}
            \big\{ \ x \in \mathbb{R}:
            \ {min} \left( X_{*j} \right)
            \ \leq \ x \ \leq
            \ \mu \left( X_{*j} \right) + \left( s - b' + 0.5 \right) \sigma \left( X_{*j} \right)
            \ \big\},
            \\ {if} \ \ s = 1
            \end{aligned}
            \\
            \begin{aligned}
            \big\{ \ x \in \mathbb{R}:
            \ \mu \left( X_{*j} \right) + \left( s - b' - 0.5 \right) \sigma \left( X_{*j} \right)
            \ < \ x \ \leq
            \ {max} \left( X_{*j} \right)
            \ \big\},
            \\ {if} \ \ s = b
            \end{aligned}
            \\
            \begin{aligned}
                \big\{ \ x \in \mathbb{R}:
                \ \mu \left( X_{*j} \right) + \left( s - b' - 0.5 \right) \sigma \left( X_{*j} \right)
                \ < \ x 
                \\
                \ \wedge \
                \ x \ \leq
                \ \mu \left( X_{*j} \right) + \left( s - b' + 0.5 \right) \sigma \left( X_{*j} \right)
                \\
                \ \wedge \
                \ {min} \left( X_{*j} \right)
                \ \leq \ x \ \leq
                \ {max} \left( X_{*j} \right)
                \ \big\},
                \\ {if} \ \ 1 < s < b
            \end{aligned}
        \end{cases}
    \end{aligned}
\end{equation}

This definition ensures that the representation of a given distribution of values is consistent across all equivalent features, regardless of the scale of their values or the interval size between minimum and maximum values. The subdomain corresponding to the center bin always contains values in the center of the distribution, between \begin{math} \mu \left( X_{*j} \right) - 0.5 \sigma \left( X_{*j} \right) \end{math} and \begin{math} \mu \left( X_{*j} \right) + 0.5 \sigma \left( X_{*j} \right) \end{math}, and is always represented by the same number, \begin{math} b' \end{math}. The other subdomains are represented by bin numbers lower or higher than it, from 1 to \begin{math} b \end{math}.

\begin{figure}[h!]
    \centering
    \includegraphics[width=\textwidth]{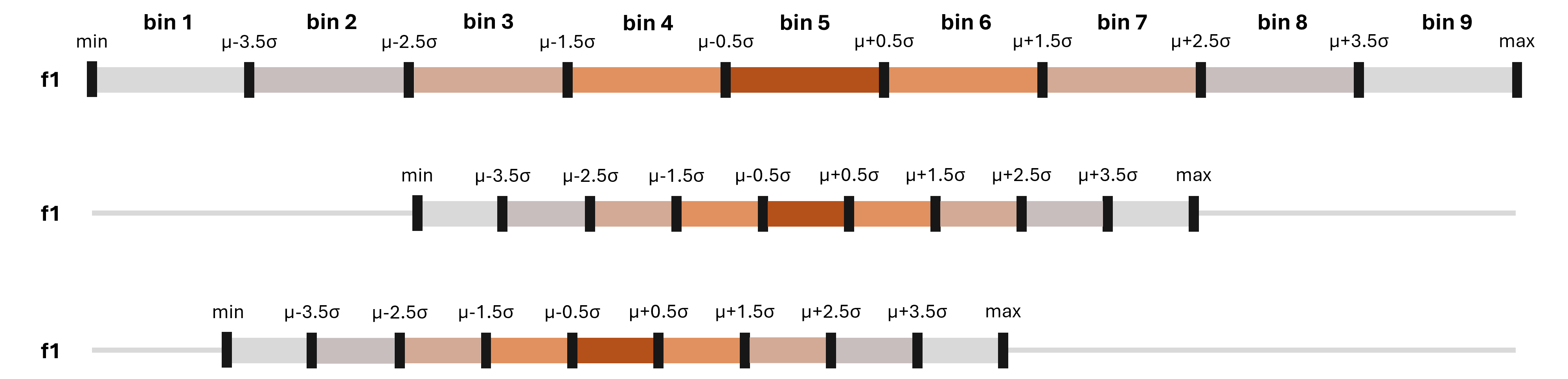}
    \caption{Subdomains of features with an equivalent distribution of values.}
    \label{fig:3-1}
\end{figure}

The consistency of the discretization is showcased with \begin{math} b = 9 \end{math} in Fig.~\ref{fig:3-1}, where three vectors with different scales and intervals actually represent the same feature f1 because they follow the same distribution of values. These three vectors are discretized into the same bins, which are highlighted in decreasing color saturation starting from the center bin.

Furthermore, this definition ensures that the representation is dynamically adapted to fit different distributions of values. The subdomains always have a size of one standard deviation, which enables the corresponding bins to vary for different features, making them compatible with each other. The exception is when a subdomain corresponds to an edge bin, the last possible bin before either the minimum or maximum values of a feature are reached. The two edge bins of a feature vector may not have a size of one standard deviation, to ensure that they always comply with the domain of a feature.

\begin{figure}[h!]
    \centering
    \includegraphics[width=\textwidth]{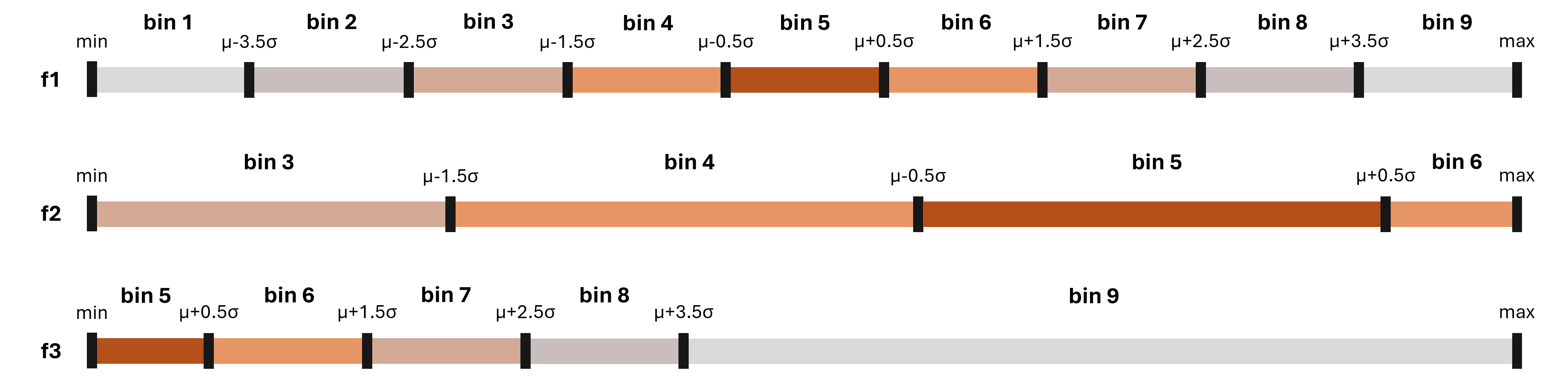}
    \caption{Subdomains of features with different distributions of values.}
    \label{fig:3-2}
\end{figure}

The dynamic size is showcased in Fig.~\ref{fig:3-2}, where three different vectors represent features f1, f2, and f3 with different distributions of values. Due to the large value of one standard deviation of f2, its minimum and maximum values were reached with just four bins, resulting in edge bins smaller than the center bin. On the other hand, since the mean of f3 is very close to its minimum, its center bin is simultaneously a small edge bin. This leads to the opposite edge bin representing a much larger interval, as f3 has fewer values in that interval. Even though these features are discretized into bins of very different sizes, their data distribution can be analyzed and compared because the same bin number is always used for equivalent subdomains.

\subsection{Recursive Projections}

Since the subdomains of a feature represent its variability in a consistent and dynamic manner, their bin numbers can be used independently, instead of the values of the original domain. Therefore, it is possible to perform a projection of any given real value to a specific bin number of any feature, creating a discrete domain-independent space with at most \begin{math} b \end{math} bin numbers.

Let \begin{math} {map} \end{math} be the function that given a real value \begin{math} x \end{math}, a feature vector \begin{math} X_{*j} \end{math}, and the maximum number of bins \begin{math} b \end{math}, maps \begin{math} x \end{math} to the bin number of its respective subdomain of \begin{math} X_{*j} \end{math}. In case that \begin{math} x \end{math} is not in the domain of \begin{math} X_{*j} \end{math}, when a value is analyzed after a dataset was already projected, it is mapped to 0 to be disregarded. This function is defined as:

\begin{equation}
    {map} \left( x, X_{*j}, b \right) \ = \
    \begin{cases}
        \ 1, \ \ {if} \ \ x \in D^1 \left( X_{*j}, b \right)
        \\
        \ \ldots
        \\
        \ b, \ \ {if} \ \ x \in D^b \left( X_{*j}, b \right)
        \\
        \ 0, \ \ {otherwise}
    \end{cases}
\end{equation}

By performing the mapping in a recursive manner, further subdomains can be identified with a higher level of granularity. The \begin{math} {map} \end{math} function can be applied to a complete feature vector to obtain the bin number of a first-level subdomain, and it can also be applied to the corresponding subvector to obtain the bin number of a second-level subdomain. These hierarchical bin numbers can be used to create a single code that represents a more granular subdomain.

Let \begin{math} {Rmap} \end{math} be the function that given a real value \begin{math} x \end{math}, a feature vector \begin{math} X_{*j} \end{math}, and the maximum number of bins \begin{math} b \end{math}, maps \begin{math} x \end{math} to the code composed of the concatenation of all the individual mappings of \begin{math} x \end{math} to the bin number of each recursive subdomain of \begin{math} X_{*j} \end{math}. The main stopping condition for the recursion occurs when the vector provided at \begin{math} X_{*j} \end{math} cannot be divided into further subdomains, as there would only be a single bin containing the entire domain of the vector. In case that \begin{math} x \end{math} is not in the domain of \begin{math} X_{*j} \end{math}, it is still mapped to 0.

In addition to the notation of curly brackets to denote unordered sets of unique elements, square brackets are used to denote that feature vectors are ordered and can have repeated elements. The \begin{math} {Rmap} \end{math} function is defined as:

\begin{equation}
    \begin{aligned}
        &{Rmap} \left( x, X_{*j}, b \right) \ = \
        \\[1.2ex]
        &\begin{cases}
            \begin{aligned}
                &\ {map} \left( x, X_{*j}, b \right),
                \\
                &\ \ \ \ \ \ \ \ \ \ \ {if} \ \ {map} \left( x, X_{*j}, b \right) \neq 0
                \ \wedge \ D^{{map} \left( x, X_{*j}, b \right)} \left( X_{*j}, b \right) = D \left( X_{*j}, b \right)
            \end{aligned}
            \\[7pt]
            \begin{aligned}
                &\ {map} \left( x, X_{*j}, b \right)
                \mathbin\Vert
                {Rmap} \big( x, \
                \big[
                a \in X_{*j}: a \in D^{{map} \left( x, X_{*j}, b \right)} ( X_{*j}, b )
                \big],
                \ b \big),
                \\
                &\ \ \ \ \ \ \ \ \ \ \ {if} \ \ {map} \left( x, X_{*j}, b \right) \neq 0
                \ \wedge \ D^{{map} \left( x, X_{*j}, b \right)} \left( X_{*j}, b \right) \subset D \left( X_{*j}, b \right)
            \end{aligned}
            \\[7pt]
            \ 0, \ \ {otherwise}
        \end{cases}
    \end{aligned}
\end{equation}

\begin{figure}[h!]
    \centering
    \includegraphics[width=\textwidth]{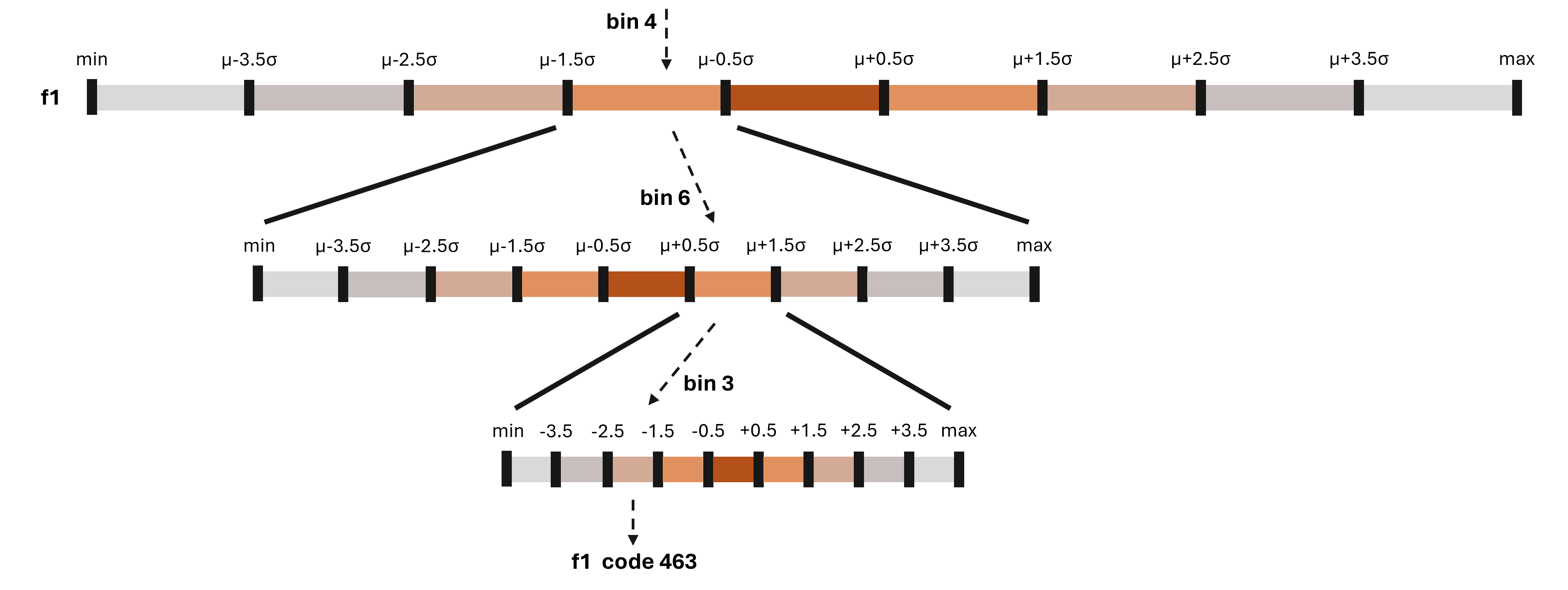}
    \caption{Recursive mapping of feature subdomains to create a code.}
    \label{fig:3-3}
\end{figure}

The recursive approach enables the domain-independent space to be expanded in a hierarchical manner to more precisely distinguish between values that are close enough to belong to the same first-level subdomain, but are different enough to be separated in a more granular subdomain. This is showcased with \begin{math} b = 9 \end{math} in Fig.~\ref{fig:3-3}, where a given value is mapped to bin number 4 of the feature vector, then to bin number 6 of that subvector, and finally to bin number 3 of that smaller subvector. Considering that the last subvector cannot be further divided, the resulting code is 463, combining those hierarchical bin numbers. In this example, with 3 levels of granularity, the discrete space can have codes from 111 to 999 to provide a more detailed representation.

By interconnecting the individual projections of a data instance \begin{math} X_{i*} \end{math} across all the features, it is possible to create tuples of codes that represent the specific combination of feature subdomains that more precisely represents \begin{math} X_{i*} \end{math}. Therefore, the projected data instances of a dataset are considered as its combinations. Like a code itself, each combination can be repeated in other data instances that have very close values, which enables the identification of similar data instances to be performed directly in the domain-independent space.

\begin{figure}[h!]
    \centering
    \includegraphics[width=\textwidth]{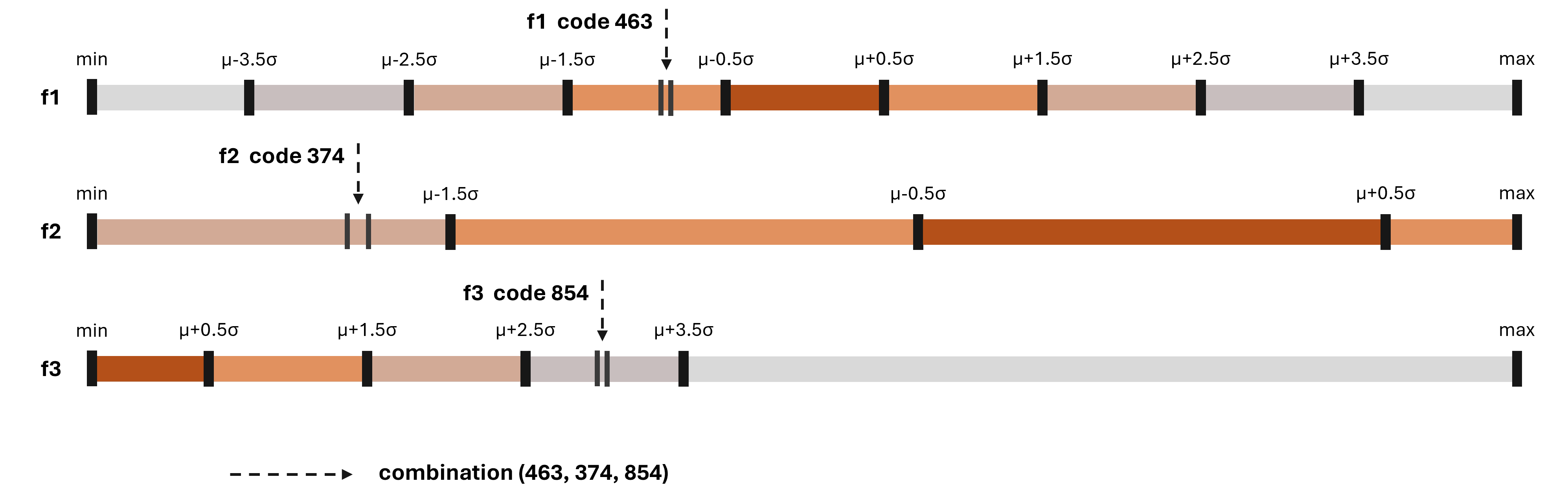}
    \caption{Parallel mapping of multiple features to create a combination.}
    \label{fig:3-4}
\end{figure}

Even though different features can have fewer subdomains and consequently smaller discrete spaces, it is ensured that those spaces are compatible with each other because they rely on the same bin numbers for equivalent subdomains. A combination is showcased in Fig.~\ref{fig:3-4}, where a data instance with features f1, f2, and f3 is mapped to codes 463, 374, and 854, respectively. The resulting projected data instance is an ordered tuple that represents that unique combination within this dataset, but does not disclose the real values of the original domain, maintaining data privacy and confidentiality.

\subsection{Projection Analysis}

The specific combinations of a dataset serve as a systematization of the underlying patterns and relationships of its features and data instances, providing useful information for the creation of documentation in the form of a data card.

The projection of all values of a dataset \begin{math} X \end{math} into a domain-independent space results in the projected dataset \begin{math} P \end{math}. Given a label vector \begin{math} y \end{math} with \begin{math} n \end{math} elements, the data instance \begin{math} X_{i*} \end{math} and its corresponding combination \begin{math} P_{i*} \end{math} are both assigned to the same class label \begin{math} y_i \end{math}, considering \begin{math} 0 \leq i < n \end{math}. Using the \begin{math} \# \end{math} symbol to denote the number of elements in a set or in a vector, the following variables can be defined to provide an overview of a projected dataset:

\begin{itemize}
    \setlength\itemsep{5pt}
    
    \item \begin{math} P^k \end{math} is the vector with all combinations in a class \begin{math} k \end{math}, i.e., the rows of the projected dataset \begin{math} P \end{math} that are assigned to a class label \begin{math} k \end{math}.

    It is expressed as: \begin{math} \ P^k = \big[ P_{i*}: y_i = k \big] \end{math}
    
    \item \begin{math} U^k \end{math} is the set with the unique combinations in a class \begin{math} k \end{math}, i.e., the unique rows of the projected subset \begin{math} P^k \end{math}.
    
    It is expressed as: \begin{math} \ U^k = \big\{ P^k \big\} \end{math}
    
    \item \begin{math} {nCode} \end{math} is the number of occurrences of a code \begin{math} p \end{math} of a feature \begin{math} j \end{math} in a class \begin{math} k \end{math}.
    
    It is expressed as: \begin{math} \ {nCode} = \# \big[ r \in \mathbb{N}_0 : P^k_{rj} = p \big] \end{math}
    
    \item \begin{math} {nCodeOverlaps} \end{math} is the number of classes that have the same code \begin{math} p \end{math} of a feature \begin{math} j \end{math}.
    
    It is expressed as: \begin{math} \ {nCodeOverlaps} = \# \big\{ s \in \mathbb{N}_0 : p \in U^s_{*j} \big\} \end{math}
    
    \item \begin{math} {nCombo} \end{math} is the number of occurrences of a unique combination \begin{math} i \end{math}, which contains a code \begin{math} p \end{math} of a feature \begin{math} j \end{math}, in a class \begin{math} k \end{math}.
    
    It is expressed as: \begin{math} \ {nCombo} = \# \big[ r \in \mathbb{N}_0 : U^k_{ij} = p \ \wedge \ U^k_{i*} = P^k_{r*} \big] \end{math}
    
    \item \begin{math} {nComboOverlaps} \end{math} is the number of classes that have the same unique combination \begin{math} i \end{math}, which contains a code \begin{math} p \end{math} of a feature \begin{math} j \end{math}.
    
    It is expressed as: \begin{math} \ {nComboOverlaps} = \# \big\{ s \in \mathbb{N}_0 : U^k_{ij} = p \ \wedge \ U^k_{i*} \in U^s \big\} \end{math}
\end{itemize}

\begin{figure}[h!]
    \centering
    \includegraphics[width=\textwidth]{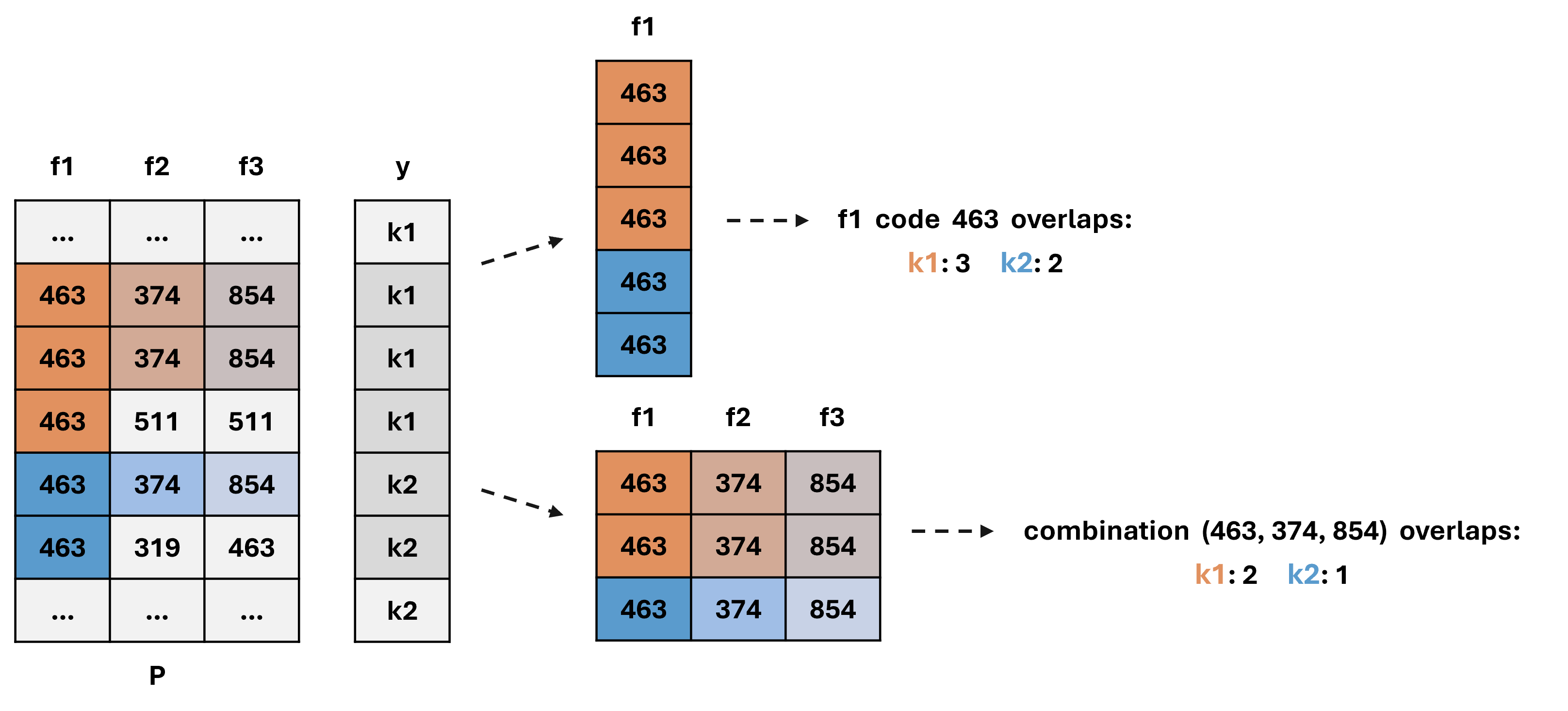}
    \caption{Overlaps of feature codes and combinations.}
    \label{fig:3-5}
\end{figure}

A simple example of code and combination overlaps is showcased with \begin{math} b = 9 \end{math} in Fig.~\ref{fig:3-5}, with the projections of classes k1 and k2 highlighted in orange and blue, respectively. This example considers only a single combination and 3 levels of granularity for simplicity, resulting in codes with 3 bin numbers. In more complex datasets, a higher level of granularity would provide a better distinction between multiple classes, with the systematization of more detailed patterns.

Overall, a SPATA pattern is the result of the projection and systematization of a tabular dataset, providing detailed and transparent information about the distributions of values and relationships of its features and data instances, without disclosing any original data instance. With SPATA patterns, datasets can be analyzed and compared, and adversarial examples and explanations of ML robustness can be generated directly in a domain-independent space.

%% file: section4.tex
\section{Experimental Validation}
\label{sec:section4}

To assess the reliability of SPATA with complex real-world datasets, an experimental validation was performed with standard benchmark datasets of the cybersecurity field. ML classification models were trained with both the original datasets and their projected versions, and their generalization and robustness were compared in a regular evaluation and an adversarial evaluation, relying on adversarial attacks. Finally, feature importance was analyzed to ensure that the impact of each feature was similar in the original and projected versions.

The following subsection describes the experimental setup, including the utilized datasets, ML models, adversarial attacks and explainability technique. Afterwards, the obtained results are presented and discussed, analyzing the differences between the original and projected datasets.

\subsection{Experimental Setup}

The experimental validation was carried out on a relatively lightweight machine equipped with a 6-core central processing unit, a 6-gigabyte graphics processing unit, and 16 gigabytes of random access memory. The following Python 3 libraries were utilized: \textit{numpy} and \textit{pandas} for general data processing, \textit{scikit-learn} for cross-validation and hyperparameter optimization, and \textit{xgboost}, \textit{lightgbm}, and \textit{tensorflow} for the implementations of the ML models.

Considering that Network Intrusion Detection (NID) is an important multi-class classification task that requires the monitoring of private network activity, two standard benchmark datasets were selected: CICIDS~\cite{Sharafaldin2018} and IoT23~\cite{Garcia2021}. The former contains complex network traffic flows of benign activity and cyber-attacks performed on a computer network, whereas the latter is focused on cyber-attacks targeting internet of things devices.

To prepare the data, one-hot encoding was employed for the conversion of categorical features. Due to their high cardinality, the low frequency categories were aggregated to avoid encoding qualitative values that had a small relevance. Finally, each dataset was randomly split into training and testing sets with 70\% and 30\% of the data instances. Since both datasets are imbalanced, the split was done with stratification to preserve the imbalanced class proportions. The resulting CICIDS dataset had 77 features, 65 numerical and 12 categorical, and the IoT23 dataset had 24 features, 4 numerical and 20 categorical.

These datasets were used to train and evalute three classification models that are commonly used for NID tasks: Extreme Gradient Boosting (XGB), Light Gradient Boosting Machine (LGBM), and Multilayer Perceptron (MLP). The first two are decision tree ensembles that rely on gradient boosting, whereas the third is a feedforward artificial neural network.

The models were fine-tuned through a grid search of typical hyperparameters for cyber-attack classification. To find the optimal configuration for each model and each dataset, a 5-fold cross-validation was performed, with 5 iterations where a model was trained with 4/5 of a training set and validated with the remaining 1/5. Afterwards, each model was evaluated with the holdout testing set.

The typically utilized metrics to evaluate a model's generalization and robustness in NID tasks are accuracy and the macro-averaged F1 score, which is designated as macro F1. Even though accuracy measures the proportion of correctly classified samples, its bias towards the majority classes must not be disregarded when the minority classes are particularly relevant, which is the case of the cyber-attack classes in NID tasks. To account for class imbalance, macro F1 computes the harmonic mean of precision and recall and gives all classes the same relevance, making it the preferred metric.

To evaluate the robustness of each model, adversarial examples were generated and combined in an adversarial testing set, with the same size as the original testing set. Two adversarial evasion attacks were combined: HopSkipJump~\cite{Chen2020} and A2PM~\cite{Vitorino2022}. The attacks were targeted, creating data perturbations in data instances of the cyber-attack classes in order to cause a model to misclassify them as benign activity.

The final analysis of the impact of each feature was performed with the feature importance values computed by SHapley Additive exPlanations (SHAP)~\cite{Lundberg2017}. This explainability technique was applied to every model across both the original and projected versions of each dataset, to provide a thorough validation.

\subsection{Results and Discussion}

A total of 229.61 seconds were required for SPATA to fully analyze the CICIDS dataset with 8 levels of granularity, systematizing all its underlying patterns and creating the projected version. Even though this dataset has over a million data instances and a large quantity of numerical features with high variability, it was analyzed relatively quickly. The time consumption was even lower on the IoT32 dataset, requiring only 9.92 seconds due to its higher proportion of categorical features, which have a simpler distribution of values.

\begin{figure}[h!]
    \centering
    \begin{minipage}[b]{0.49\textwidth}
        \includegraphics[width=\textwidth]{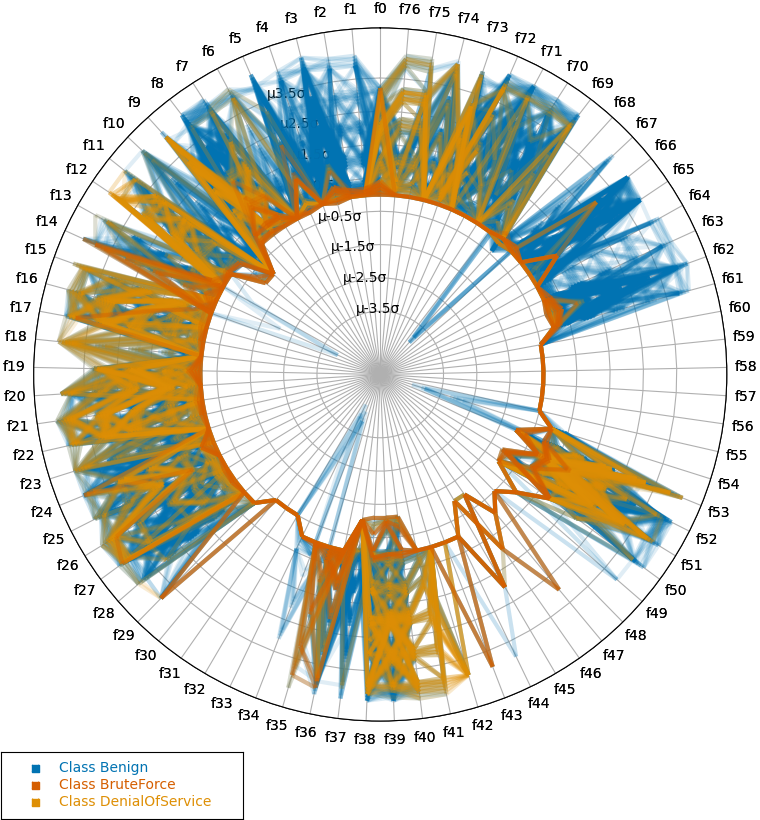}
    \end{minipage}
    \hfill
    \begin{minipage}[b]{0.49\textwidth}
        \includegraphics[width=\textwidth]{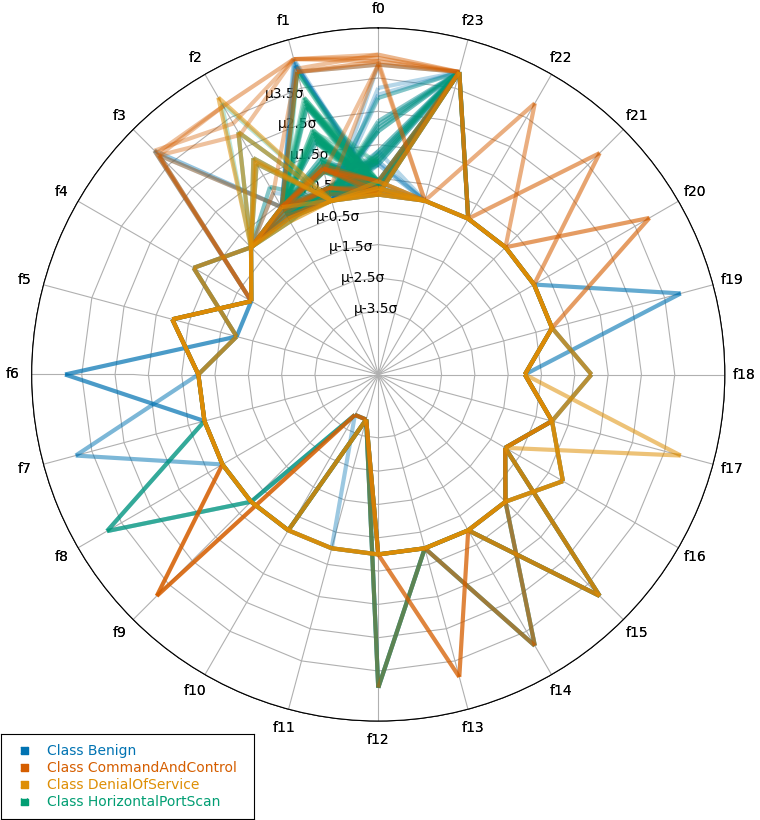}
    \end{minipage}
\caption{SPATA patterns of CICIDS dataset (left) and IoT23 dataset (right).}
\label{fig:4-pat}
\end{figure}

The graphical overview of the SPATA patterns of CICIDS and IoT23 is presented in Fig.~\ref{fig:4-pat}. Each class is plotted in a different color, and less opaque segments indicate less frequent feature combinations. Despite being just an overview, it can be observed that the patterns enable the identification of the parts of the discrete space where each class is clearly distinguishable, as well as the parts where there is an overlap between multiple classes, both in CICIDS and IoT23, which has fewer combinations.

After creating the projections, it was important to validate that they were suitable replacements of the original data for ML robustness assessments. To perform this validation, distinct ML models were trained with the original and projected datasets, and their generalization and robustness were compared considering accuracy and macro F1. Table~\ref{tab:results_cicids} presents the obtained results for the two versions of the CICIDS dataset, with the "Attack" column indicating whether a score was before (cross) or after (checkmark) an adversarial evasion attack.

\begin{table}[h!]
\centering
\caption{Evaluation results for CICIDS dataset.}
\label{tab:results_cicids}
\begin{tabular}{p{0.12\textwidth} p{0.14\textwidth} p{0.12\textwidth} p{0.22\textwidth} p{0.22\textwidth}}
\toprule
\textbf{Model} & \textbf{Metric} & \textbf{Attack} & \textbf{Original Score} & \textbf{SPATA Score} \\
\bottomrule
\multirow{4}{*}{XGB} & \multirow{2}{*}{Accuracy} & \tablespace\xmark & 99.97 & 99.97 \\
                     &                           & \tablespace\cmark & 84.29 & 84.11 \\
                     \cline{2-5}
                     & \multirow{2}{*}{Macro F1} & \tablespace\xmark & 99.92 & 99.92 \\
                     &                           & \tablespace\cmark & 34.03 & 30.94 \\
\hline
\multirow{4}{*}{LGBM} & \multirow{2}{*}{Accuracy} & \tablespace\xmark & 99.97 & 99.97 \\
                     &                           & \tablespace\cmark & 85.10 & 84.26 \\
                     \cline{2-5}
                     & \multirow{2}{*}{Macro F1} & \tablespace\xmark & 99.91 & 99.91 \\
                     &                           & \tablespace\cmark & 35.62 & 31.58 \\
\hline
\multirow{4}{*}{MLP} & \multirow{2}{*}{Accuracy} & \tablespace\xmark & 97.06 & \textbf{97.65} \\
                     &                           & \tablespace\cmark & 85.52 & \textbf{93.04} \\
                     \cline{2-5}
                     & \multirow{2}{*}{Macro F1} & \tablespace\xmark & 92.61 & \textbf{98.59} \\
                     &                           & \tablespace\cmark & 53.74 & \textbf{74.07} \\
\hline
\end{tabular}
\end{table}

Even though the XGB and LGBM models created with the original CICIDS training set obtained very high scores, with up to 99.97\% accuracy and 99.92\% macro F1, the models created with the projected training set were able to achieve the same scores. Since the score differences were just in the fourth decimal place, only a few additional predictions of these models became incorrect. Therefore, the decision tree ensembles preserved an equivalent generalization and exhibited a similar behavior in the projected dataset.

On the other hand, MLP stood out for achieving slightly better results with the projected dataset. The original model could not detect all the data variations of the minority brute force class and only reached a macro F1 of 92.61\%, despite hyperparameter optimization attempts that included up to three hidden layers. This score was surpassed by the model trained with the projected data, which improved almost 6\% with the same hyperparameters, reaching 98.59\%. This suggests that training an artificial neural network with the projected data may sometimes help it to better distinguish between minority classes.

When adversarial evasion attacks were performed against the three original models, their scores were significantly lower, as they are vulnerable to adversarial examples. Performing the same attacks against the models trained with the projected data resulted in an equivalent robustness, with the macro F1 of XGB and LGBM being only 4\% and 3\% lower. The exception was again MLP, which exhibited a better robustness, with a macro F1 approximately 20\% higher.

The same regular and adversarial evaluations were repeated with the IoT23 dataset. Table~\ref{tab:results_iot23} presents the obtained results for the two versions of this dataset.

\begin{table}[h!]
\centering
\caption{Evaluation results for IoT23 dataset.}
\label{tab:results_iot23}
\begin{tabular}{p{0.12\textwidth} p{0.14\textwidth} p{0.12\textwidth} p{0.22\textwidth} p{0.22\textwidth}}
\toprule
\textbf{Model} & \textbf{Metric} & \textbf{Attack} & \textbf{Original Score} & \textbf{SPATA Score} \\
\bottomrule
\multirow{4}{*}{XGB} & \multirow{2}{*}{Accuracy} & \tablespace\xmark & 95.54 & 95.49 \\
                     &                           & \tablespace\cmark & 46.24 & \textbf{59.98} \\
                     \cline{2-5}
                     & \multirow{2}{*}{Macro F1} & \tablespace\xmark & 92.05 & 91.69 \\
                     &                           & \tablespace\cmark & 36.13 & \textbf{45.73} \\
\hline
\multirow{4}{*}{LGBM} & \multirow{2}{*}{Accuracy} & \tablespace\xmark & 95.55 & 95.51 \\
                     &                           & \tablespace\cmark & 53.10 & \textbf{61.70} \\
                     \cline{2-5}
                     & \multirow{2}{*}{Macro F1} & \tablespace\xmark & 92.03 & 91.96 \\
                     &                           & \tablespace\cmark & 41.75 & \textbf{43.74} \\
\hline
\multirow{4}{*}{MLP} & \multirow{2}{*}{Accuracy} & \tablespace\xmark & 95.29 & 93.64 \\
                     &                           & \tablespace\cmark & 51.95 & 50.86 \\
                     \cline{2-5}
                     & \multirow{2}{*}{Macro F1} & \tablespace\xmark & 82.63 & 78.34 \\
                     &                           & \tablespace\cmark & 33.95 & 26.68 \\
\hline
\end{tabular}
\end{table}

The models trained with the projected IoT23 training set were not able to preserve the same scores as in the original version, possibly due to the smaller data variability of this dataset, which makes it more difficult to distinguish between the different classes in the projected space. The smallest decreases were achieved by LGBM, with just -0.04\% accuracy and -0.07\% macro F1, whereas the largest were obtained by MLP, with -1.65\% accuracy and -4.29\% macro F1. Nevertheless, since these are relatively small decreases, all models generalized well and there were no substantial differences in their behavior.

In this dataset, the XGB and LGBM models trained with the projected data exhibited a better robustness than with the original version. The adversarial evasion attacks could not cause as many misclassifications, resulting in accuracy and macro F1 up to approximately 14\% and 10\% higher. Despite these differences, the models maintained a relatively similar robustness, as the moderately better scores can be caused by the pseudo-random number generation that is part of each adversarial attack iteration.

Finally, the impact of the projected dataset on the behavior of the ML models was analyzed in more detail, taking into account feature importance. By applying the SHAP explainability technique to every model across both versions of each dataset, it was possible to analyze whether the most impactful features of the original versions remained as impactful in the projected versions.

To summarize the performed analysis, Table~\ref{tab:results_features} presents the ranking differences of the top 10 features of each dataset. Considering the original rank of a given feature as its baseline, each column from "1st" to "10th" indicates whether the rank of that feature was improved or worsened in the projected version. For instance, since the 10th most impactful feature for the XGB model in the original CICIDS dataset became only the 13th most impactful in its projected version, the column "10th" contains the value "-3".

\begin{table}[h!]
\centering
\caption{Top 10 feature ranking differences from original to projected datasets.}
\label{tab:results_features}
\begin{tabular}{p{0.13\textwidth} p{0.13\textwidth} p{0.06\textwidth} p{0.06\textwidth} p{0.06\textwidth} p{0.06\textwidth} p{0.06\textwidth} p{0.06\textwidth} p{0.06\textwidth} p{0.06\textwidth} p{0.06\textwidth} p{0.06\textwidth}}
\toprule
\textbf{Dataset} & \textbf{Model} & \textbf{1st} & \textbf{2nd} & \textbf{3rd} & \textbf{4th} & \textbf{5th} & \textbf{6th} & \textbf{7th} & \textbf{8th} & \textbf{9th} & \textbf{10th} \\
\bottomrule
\multirow{3}{*}{CICIDS} & XGB & $=$ & $=$ & -9 & $=$ & $=$ & -14 & +1 & -17 & +2 & -3 \\
                     & LGBM & $=$ & $=$ & -16 & -22 & -4 & +2 & -1 & -11 & +6 & -1 \\
                     & MLP & $=$ & -12 & -24 & -4 & -8 & -22 & -17 & +1 & -2 & -16 \\
\hline
\multirow{3}{*}{IoT23} & XGB & $=$ & $=$ & $=$ & $=$ & -6 & +1 & +1 & $=$ & $=$ & +3 \\
                     & LGBM & $=$ & $=$ & $=$ & -1 & +1 & $=$ & -1 & -1 & +2 & -2 \\
                     & MLP & $=$ & $=$ & -5 & -8 & -8 & $=$ & $=$ & +6 & +6 & -4 \\
\hline
\end{tabular}
\end{table}

Despite some changes in the ranking of each dataset's features, it can be observed that most of the originally top 10 features remained in the top 10 in the projected datasets. These features continued to be impactful, which enables the behavior of an ML model to be analyzed and explained with the projected version without the need to directly access the original data.

However, it is important to note that the decision boundaries of the three considered ML models did not remain exactly the same. Since all feature vectors were projected to the same discrete space, some small amount of loss of information was expected to occur. For instance, the 3rd most important feature for the original LGBM for CICIDS was \textit{bwd-packet-length-std}, indicating the deviation of the size of response packets in a network connection. This is typically an important feature for NID tasks, but its importance was decreased in the projected version, which may misrepresent LGBM's behavior.

Nevertheless, some changes to the decision boundaries seem beneficial. In the case of MLP for CICIDS, the two most impactful features were \textit{flow-iat-mean} and \textit{flow-iat-std}, indicating the mean and deviation in the inter-arrival time between different packets. The second feature was replaced by \textit{flow-duration}, which was missing only in the original MLP's top 10. Since this feature is typically useful for the detection of the Denial-of-Service class of attacks, its increased importance helps explain why the evaluation scores of MLP were improved.

Overall, the obtained results demonstrate the reliability of SPATA for the projection of tabular datasets into a domain-independent space without significant loss of information. The SPATA patterns enabled the training of ML models that achieved a similar generalization and robustness, and exhibited feature importances that were relatively similar to those of the original NID datasets, reducing the need to directly access and analyze private datasets.

%% file: section5.tex
\section{Conclusions}
\label{sec:section5}

This paper tackled the challenge of creating more detailed and transparent data cards, to enable ML robustness assessments to be performed and explainability techniques to be applied while maintaining data privacy and confidentiality. The SPATA method was presented and validated, creating domain-independent representations of the statistical patterns of tabular datasets to eliminate the need to use the original data. The obtained results demonstrate the reliability of SPATA and its applicability to complex real-world tabular datasets.

Since the analysis of a dataset is being performed directly in a domain-independent space, it paves the way for more complex analyses and the generation of interpretable explanations of data quality and ML robustness. By using the systematized SPATA patterns as a data card, it may be possible to identify correlations between multiple features and perform synthetic data generation to improve data diversity. Furthermore, by projecting the class predictions of an ML model, it may also be possible to create model cards and use them to directly generate adversarial examples, identifying how different feature combinations affect the misclassifications of a model.

In the future, it may be possible to compare the differences between the projection of a labeled dataset and the projection of a model's class predictions, identifying vulnerable spaces in its decision boundaries and improving interpretability. There is ongoing research and development work to expand SPATA and explore approaches to perform these analyses directly in the domain-independent space, contributing to the evaluation of different types of ML models across different datasets and to the development of more trustworthy AI.